%% file: main.tex
\documentclass[11pt]{article}

\usepackage{naacl2021}

\usepackage{times}
\usepackage{url}
\usepackage{latexsym}

\usepackage[T1]{fontenc}

\usepackage[utf8]{inputenc}

\usepackage{microtype}

\usepackage{amsmath}

\usepackage{xcolor}
\usepackage{graphicx}
\usepackage{multirow}
\usepackage{stfloats}

\title{Noise Robust Named Entity Understanding for Voice Assistants}

\author{
  Deepak Muralidharan\thanks{ \textbf{ \ Equal contributions.}},\enspace Joel Ruben Antony Moniz\footnote[1]{},\enspace Sida Gao\footnote[1]{},\enspace Xiao Yang\footnote[1]{} \thanks{ \enspace graceyx.scut@gmail.com}, \\ \textbf{Justine Kao,\enspace Stephen Pulman,\enspace Atish Kothari,\enspace Ray Shen,\enspace Yinying Pan,} \\ \textbf{Vivek Kaul, \enspace Mubarak Seyed Ibrahim,\enspace Gang Xiang,\enspace Nan Dun,\enspace Yidan Zhou,}\\ \textbf{Andy O,\enspace Yuan Zhang,\enspace Pooja Chitkara,\enspace Xuan Wang,\enspace Alkesh Patel,}\\ \textbf{Kushal Tayal,\enspace Roger Zheng, \enspace Peter Grasch,\enspace Jason D. Williams,\enspace Lin Li \thanks{ \enspace lli9@apple.com} } \\
  Apple}

\date{}

\begin{document}
\maketitle
\begin{abstract}
	\input{sections/s0-abstract.tex}
\end{abstract}

\input{sections/s1-intro.tex}

\input{sections/s2-lit_review.tex}
\input{sections/s3-method.tex}

\input{sections/s4-expt_setup.tex}
\input{sections/s5-results.tex}

\input{sections/s6-conclusion.tex}
\input{sections/s7-ethics.tex}

\section*{Acknowledgements}
We would like to thank Alex Acero, Anmol Walia, Arjun Rangarajan, Barry-John Theobald, Bhagyashree Shekawat, Christopher Klein, Dhivya Piraviperumal, Hong Yu, Jianpeng Cheng, Jiarui Lu, John Giannandrea,  John Keesling, Katy Linsky, Lanqing Wang, Ryan Templin, Robert Daland, Shruti Bhargava, Xi Chen and Yvonne Xiao for discussions and for their help and feedback. We also want to thank all the anonymous reviewers for the helpful feedback and suggestions on this work.

\bibliographystyle{acl_natbib}
\bibliography{biblio/biblio,biblio/biblio_deepak,biblio/biblio_joel,biblio/biblio_xiao}

\appendix
\onecolumn

\input{sections/sa-appendix.tex}

\end{document}

%% file: sections/s0-abstract.tex
Named Entity Recognition (NER) and Entity Linking (EL) play an essential role in voice assistant interaction, but are
challenging due to the special difficulties associated with spoken user queries.  In this paper, we propose a novel 
architecture that jointly solves the NER and EL tasks by 
combining them in a joint reranking module. We show that
our proposed framework improves NER accuracy by up to 3.13\% and EL accuracy by up to 3.6\% in F1 score.
The features used  also lead to better accuracies in other natural language understanding tasks, such as
domain classification and semantic parsing.

%% file: sections/s1-intro.tex
\section{Introduction}
\label{sec:intro}

Understanding named entities correctly when interacting with virtual assistants (e.g. ``Call Jon'', ``Play Adele
hello'', ``Score for Warrior Kings game'') is crucial for a satisfying user experience. However, NER and EL methods that work well
on written text often perform poorly in such applications:
utterances are relatively short (with just $5$ tokens, on average), so there is not much
context to help disambiguate; speech recognizers make errors (``Play Bohemian raspberry'' for ``Play Bohemian Rhapsody"); users also make
mistakes (``Cristiano Nando'' for ``Cristiano Ronaldo''); non-canonical forms of names are frequent (``Shaq'' for
``Shaquille O'Neal''); and users often mention new entities unknown to the system. 

In order to address these issues  we  propose a novel Named Entity Understanding (NEU) system that combines and optimizes NER and EL for noisy spoken
  natural language utterances. We pass multiple NER hypotheses to EL for reranking, enabling NER to benefit from EL by
  including information from the knowledge base (KB).
  
We also design a retrieval engine tuned for spoken utterances for retrieving candidates from the KB. The retrieval engine, along with other techniques devised to address fuzzy entity mentions, lets the EL model be more robust to partial mentions, variation in named entities, use of aliases, as well as human and speech transcription errors.

Finally,  we demonstrate that our framework can also empower other natural language understanding tasks, such as domain classification (a sentence classification task) and semantic parsing.

%% file: sections/s2-lit_review.tex
\section{Related Work}
\label{sec:rel_work}

There have been a few attempts to explore NER on the output of a speech pipeline
\cite{ghannay2018end,abujabal2018neural,coucke2018snips}. Among these, our NER model is closest to
\newcite{abujabal2018neural} and \newcite{coucke2018snips}; however, unlike the former, we use a richer set of features rather than phonemes as input, and unlike the latter, we are able to use a deep model because of the large volume of data available.

EL has been well explored in 
the context of clean \cite{martins-etal-2019-joint,kolitsas-etal-2018-end,luo2015joint} and noisy text inputs
\cite{eshel-etal-2017-named,guo-etal-2013-link,liu-etal-2013-entity}, but as with NER, there have been only a few
efforts to explore EL in the context of transcribed speech \cite{benton-dredze-2015-entity,gao-2017-support}, although crucially, both these works assume gold standard NER and focus purely on the EL component.

Traditionally, a pipelined architecture of NER followed by EL has been used to address the entity linking task
\cite{lin-etal-2012-entity,derczynski2015analysis,bontcheva2017crowdsourcing,bowden-etal-2018-slugnerds}. Since these approaches rely only on the best NER hypothesis, errors from NER propagate to the EL step.
To alleviate this, joint models have been proposed: \newcite{sil-yates} proposed an NER+EL model which re-ranks candidate mentions and entity links produced by their
base model. Our work differs in that we use a high precision NER system, while they use a large number of heuristically
obtained Noun Phrase (NP) chunks and word n-grams as input to the EL stage. \newcite{luo2015joint} jointly train an NER
and EL system using a probabilistic graphical model. However, these systems are trained and tested on clean text and do
not address the noise problems we are concerned with.

%% file: sections/s3-method.tex
\section{Architecture Design}

\label{sec:method}
 
For a given utterance, we first detect and label entities
using the NER model and generate the top-$l$ candidate hypotheses using beam search.
The EL model consists of two stages: (i) candidate retrieval and (ii) joint linking and re-ranking.
In the retrieval stage, for each NER hypothesis, we construct a structured search query
and retrieve the top-$c$ candidates from the retrieval engine.
In the ranking stage, we use a neural network to rank these
candidate entity links within each NER hypothesis while simultaneously using rich signals (entity popularity,
similarity between entity embeddings, the relation across multiple entities in one utterance, etc.) from these entity links as additional features to re-rank the
NER hypotheses from the previous step, thus jointly addressing both the NER and EL tasks.

\input{sections/s3.1-ner.tex}

\input{sections/s3.2-el.tex}
\input{sections/s3.3-application.tex}

%% file: sections/s3.1-ner.tex
\subsection{NER}
\label{sec:ner}

For the NER task, following \newcite{lample-etal-2016-neural}
we use a combination of character and word level features.
They are extracted by a bi-directional LSTM (biLSTM) \cite{hochreiter1997long}, and then concatenated with pre-trained GloVe word embeddings \footnote{We also tried more recent contextual
embeddings such as BERT \cite{devlin2019bert}, and empirically observed very little difference in performance when compared to GloVe. So we adopt GloVE, which is substantially more efficient in terms of inference time required.}
\cite{pennington-etal-2014-glove} to pass through another biLSTM and fed into a CRF model to produce
the final label prediction based on a score $s( \mathbf{\tilde{y_{i}}}, \mathbf{x}; \mathbf{\theta} )$ that jointly optimizes the
probability of labels for the tokens and the transition score for the entire sequence
$\mathbf{\tilde{y_{i}}}=(y_1,\dots,y_T)$ given the input $\mathbf{x}$:
\begin{equation*}
  s( \mathbf{\tilde{y_{i}}}, \mathbf{x}; \mathbf{\theta} ) = \sum_{t=0}^{T} \left( \psi_{t, \mathbf \theta}(y_t) +
    \phi_{t,t+1} (y_t, y_{t+1}) \right),
\end{equation*}

\noindent where $\psi_{t,\theta}$ is the biLSTM prediction score from the label $y_t$ of the $t$\textsuperscript{th} token, and $\phi(j, k)$ is the transition score 
from label $j$ to label $k$.

During training, we maximize the probability of the correct label sequence $p_{seq}$, which is defined as
\begin{equation*}
p_{seq} ( \mathbf{\tilde{y_{i}}}, \mathbf{x};  \mathbf  \theta ) = \frac{\exp({s( \mathbf{\tilde{y_{i}}}, \mathbf{x};  \mathbf  \theta )) }}{\sum_{\mathbf{\tilde{y_{j}}} \in S} \exp{(s( \mathbf{\tilde{y_{j}}}, \mathbf{x};  \mathbf  \theta ))}},
\end{equation*}

\noindent where $\mathbf{\tilde{y_{i}}}$ is the label sequence for hypothesis $i$, and $S$ is the set of all possible label sequences. 

During inference, we generate up to 5 NER alternatives for each utterance using beam search.
We also calculate a mention level confidence $p_{men}$ for each 
entity mention $ \mathbf{m_k}$.  $p_{men}$ is computed by aggregating the sequence level confidence for all the prediction sequences that
share the same mention sub-path $\mathbf{m_k}$:
\begin{equation*}
p_{men} (\mathbf{m_k}, \mathbf x ; \mathbf \theta) = \frac{ \sum_{ \mathbf{\tilde{y_{i}}} \in S_{\mathbf{m_i}}} \exp({s (\mathbf{\tilde{y_{i}}}, \mathbf{x};  \mathbf  \theta )}) }{\sum_{ \mathbf{\tilde{y_j}} \in S} \exp(s (\mathbf{\tilde{y_{j}}}, \mathbf{x};  \mathbf  \theta ))},
\end{equation*}

\noindent where ${S_{m_i}}$ is the set of prediction sequences that all have $ \mathbf{m_k}$ as the prediction for the corresponding tokens. 
Both $p_{seq}$ and $p_{men}$ are computed by dynamic programming, and serve as informative features in the EL model.

%% file: sections/s3.2-el.tex
\subsection{Joint Linking and Re-ranking}
\label{subsec:el}

\begin{table*}[b]
  \begin{equation}
    s(\mathbf{y}) = \textbf{MLP} (
\mathbf{f}_{\text{utter}} \oplus \mathbf{f}_{\text{NER}} \oplus \mathbf{f}_{\text{CR}}
 \bigoplus_{j=1}^{k} [\textbf{MLP}(\textbf{CNN}(\mathbf{m}_j), \textbf{CNN}(\mathbf{e}_j)) \oplus \\  \textbf{CNN}(\mathbf{m}_j) \oplus \textbf{CNN}(\mathbf{e}_j)]
 )
 \label{eq-nen-score}
\end{equation}
\end{table*}

The entity linking system follows the NER model and consists of two steps: candidate retrieval,
and joint linking and re-ranking. 

To build the candidate retrieval engine, we first index the list of entities in our knowledge base, which can be updated daily to capture new entities and change of their popularity. 
To construct the index, we iterate through the flattened list of entities and construct
token-level unigram, bigram and trigram terms from the surface form of each entity.
Apart from using the original entity names, we also use common aliases, harvested from usage logs, for popular entities (e.g. LOTR as an alias for ``Lord of the Rings'') to make the retrieval engine more
robust to commonly occurring variations.
Next, we create an inverted index which maps the unique list of n-gram terms to the list of entities that these n-grams are part of, also known as posting lists.
Further, to capture cross-entity relationships in the knowledge base (such as relationships between an artist and a song or two sports teams belonging to the same league),
we assign a pointer\footnote{Each entity in our knowledge base consists of metadata (for example, a song entry in our knowledge base would contain metadata such as the music artist, album, year the song was released in etc.) that we leverage to automatically construct these relationship pointers.} for each entity in the knowledge base to its related entities and this relational information is leveraged by the EL model for entity disambiguation (described in \ref{sec:qualitative}).
We then compute the tf-idf score 
for all the n-gram terms present in the entities and store them in the inverted index.

For each hypothesis predicted by the NER model we query the retrieval engine with the corresponding
text.
We first send the query through a high-precision seq-to-seq correction model \cite{schmaltz-etal-2017-adapting,ge-etal-2019-automatic}
trained using common errors observed in usage. 
Next, we construct n-gram features from the corrected query in a similar way to the indexing phase and
retrieve all entities matching these n-gram features in our inverted index.
Additionally, we use synonyms derived from usage for each term in the query to expand our
search criteria: for example, our synonym list for ``Friend" contains ``Friends",  which matches the TV show name which
would have been missed if only the original term was used.

For each entity retrieved, we get the tf-idf score for the terms present in the query chunk from the
inverted index.
We then aggregate the tf-idf scores of all the terms present in the query for this entity and linearly combine
this aggregate score with other attributes such as popularity (i.e. prior usage probability) of the entity to generate a final
score for all retrieved entity candidates for this query.
Finally, we perform an efficient sort across all the entity candidates based on this score and return a top-$c$
(in our case $c$ = 25) list filtered by the entity type detected by the NER model for that hypothesis.
These entity candidates coupled with the original NER hypothesis are sent to the ranker model described below 
for joint linking and re-ranking.

Following the candidate retrieval step, 
we introduce a neural model to rerank the candidate entities, aggregating features from both the NER model and the candidate retrieval engine.

The EL model scores each entity linking hypothesis separately. An entity linking hypothesis consists of a prediction from the NER model (which consists of named entity chunks in the input utterance and their types), and the candidate retrieval results for each chunk. Formally, we define an entity linking hypothesis $\mathbf{y}$ with $k$ entity predictions as:
\begin{equation*}
\mathbf{y} = \{ \mathbf{f}_{\text{utter}}, \mathbf{f}_{\text{NER}}, \mathbf{f}_{\text{CR}}, \{j \in \{1 \dots k\}:
	( \mathbf{m}_j, \mathbf{e}_j )
\} \}
\end{equation*}

\noindent where $\mathbf{m}_j$ is the $j$-th mention in the utterance, and $\mathbf{e}_j$ is the entity name associated with this mention from the knowledge base. $\mathbf{f}_{\text{utter}}$, $\mathbf{f}_{\text{NER}}$, $\mathbf{f}_{\text{CR}}$ are features derived from the original utterance text, the NER model and the candidate retrieval system respectively. In our system, $\mathbf{f}_{\text{utter}}$ is a representation of the utterance from averaging the pre-trained word embeddings for the tokens in the utterance. Intuitively, having a dense representation of the full utterance can help the EL model better leverage signals from the utterance context. $\mathbf{f}_{\text{NER}}$ includes the type of each mention, as well as the sequence and mention confidence computed by the NER model. $\mathbf{f}_{\text{CR}}$ includes popularity, and whether a relation exists between the retrieved entities in $\mathbf{y}$. 

To be robust to noise, the EL model adopts a pair of CNNs to compare each entity mention $\mathbf{m}_j$ and its
corresponding knowledge base entity name $\mathbf{e}_j$. The CNN learns a name embedding with one-dimensional
convolution on the character sequence, and the kernel parameters are shared between the CNN used for user mention and the one used for the canonical name. A character-based text representation model is better at handling mis-transcriptions or mis-pronounced entity names. While a noisy entity name may be far from the canonical name in the word embedding space when they are semantically different, they are usually close to each other in the character embedding space due to similar spellings.
To model the similarity between CNN name embeddings of $\mathbf{m}_j$ and $\mathbf{e}_j$, we use the standard cosine
similarity as a baseline, we experiment with an MLP that takes the concatenated name embeddings as input.
We are able to model more expressive interactions between the two name embeddings with the MLP, and in turn better handle errors.
Finally, we concatenate the similarity features with other features as input to another MLP that computes
the final score for $\mathbf{y}$. Formally, the scoring function is defined in Equation \ref{eq-nen-score}, where $\oplus$ means concatenation.

In our data, the number of entity mentions in an utterance averages less than 3. We pad the entity feature sequence
to length 5, which provides a good coverage. In the scoring model above, we use a simple concatenation to
aggregate the embedding similarities of multiple entity mentions which empirically performs as well as sequence models like
LSTM, while being much cheaper in computation.

To train the EL model, we use the standard max-margin loss for ranking tasks. If for the $i$-th example, we denote the ground truth as $\mathbf{y^*}_i$ and an incorrect prediction as $\mathbf{\hat{y}}_i$, and the scoring function $s(\cdot)$ is as defined in Equation \ref{eq-nen-score}, the loss function is
\begin{equation}
\mathcal{L} = \frac{1}{N}\sum_{i = 1}^{N} [\gamma(\mathbf{\hat{y}}_i, \mathbf{y^*}_i) + s(\mathbf{\hat{y}}_i) -
s(\mathbf{y^*}_i)]_+.
\label{eq-nen-loss}
\end{equation}
The max-margin loss
encourages the ground truth score to be at least a margin $\gamma$ higher than the score of an incorrect
prediction. The margin is defined as a function of the ground truth and the incorrect prediction, thus adaptive to the
quality of prediction. A larger margin is needed when the incorrect prediction is further away from the ground
truth. For our reranking task, we set a smaller margin when only the resolved entities are incorrect but the NER result
is correct, and a larger margin when the NER result is wrong. This adaptive margin helps rerank NER hypotheses even when the model cannot rank the linking results correctly. During training, we uniformly
sample the negative predictions from the candidates retrieved by the retrieval engine.

%% file: sections/s3.3-application.tex
\subsection{Improvement on Other Language Understanding Tasks} 
\label{subsec:method_application}

We also explore the impact of our NEU feature encoding on two tasks: a domain classifier and a domain-specific shallow semantic parser.

\subsubsection{Domain Classification}
\label{subsubsec:method_intent}

Domain classification identifies which domain a user's request falls into: sports, weather, music, etc., and is usually done by posing the task as sequence classification: our baseline uses word embeddings and gazetteer features as inputs to an RNN, in a manner similar to \newcite{chen2019active}.

Consider a specific token $t$. Let $a$ be the number of alternatives used from the NER model in the domain classifier
(which we treat as a hyperparameter), $p_{i}$ represent the (scalar) sequence level confidence score $p_{seq} ( \mathbf{\tilde{y_{i}}}, \mathbf{x};  \mathbf  \theta )$ of the
$i$\textsuperscript{th} NER alternative defined in Section~\ref{sec:ner}, $c_i$ represent an integer for the entity type that NER hypothesis $i$ 
assigns to the token $t$, and $\mathbf{o}(.)$ represent a function converting an integer into its corresponding one-hot
vector. Then the additional NER feature vector $\mathbf{f_{r}}$ concatenated to the input vector fed into token $t$ as part of the domain classifier can be written as:

\begin{equation} \label{eq:dc_ner}
    \mathbf{f_{r}} = \bigoplus_{i=1}^{i=a} p_i\mathbf{o}(c_i).
\end{equation}

Likewise, for the featurization that uses both NER and EL, let $a$ be the number of alternatives used from the NER+EL system in the domain classifier (also a hyperparameter); these $a$ alternatives are now sorted by the scores from the EL hypotheses, as opposed to the sequence level confidence scores from NER. Let $s_i$ be the $i$\textsuperscript{th} re-ranked alternative's cosine similarity score between the mention and knowledge base entity name as output by the EL model. $p_i$ and $c_i$ are consistent with our earlier notation, except that they now correspond to the $i$\textsuperscript{th} NER alternative after re-ranking. Then the additional NER+EL feature vector $\mathbf{f_{u}}$ concatenated to the input fed into token $t$ as part of the domain classifier can be written as:
\begin{equation} \label{eq:dc_rer}
    \mathbf{f_{u}} = \bigoplus_{i=1}^{i=a} p_i\mathbf{o}(c_i) \oplus s_i\mathbf{o}(c_i).
\end{equation}

\subsubsection{Semantic Parsing}
\label{subsubsec:method_parse}

Our virtual assistant also uses domain-specific shallow semantic parsers, running after domain classification, responsible both for identifying the correct intent that the user expects (such as the ``play'' intent associated with a song) and for assigning semantic labels to each of the tokens in a user's utterance (such as the word ``score" and ``game'' respectively being tagged as tokens related to a sports statistic and sports event respectively in the utterance ``What's the score of yesterday's Warriors game?''). Each semantic parser is structured as a multi-task sequence classification (for the intent) and sequence tagging (for the token-level semantic labelling) task, with our production baseline using word embeddings and gazetteer features as inputs into an RNN similar to our domain classifier.
Here, $\mathbf{f_{r}}$ and $\mathbf{f_{u}}$ are featurized as described above.
Note that in contrast to the NEU system, the semantic parser uses a domain-specific ontology, to enable each domain to work independently and to not be encumbered by the need to align ontologies.

%% file: sections/s4-expt_setup.tex
\section{Datasets and Training Methodology}
\label{sec:expt_setup}

To create our datasets, we randomly sampled around $600k$ unique anonymous English transcripts
(machine transcribed utterances), and annotated them with NER and EL labels.
Utterances are subject to Apple’s baseline privacy practices with respect to Siri requests, including that such requests
are not associated with a user’s Apple ID, email address, or other data Apple may have from a user’s use of other Apple
services, and have been filtered as described in Section \ref{sec:ethics}.
We then split the annotated data into 80/10/10 for train, development and test sets.
For both the NER and EL tasks, we report our results on test sets sampled from
the ``music'', ``sports'' and ``movie \& TV'' domains. These are popular domains in the usage and have a high
percentage of named entities: with an average of 0.6, 1.1 and 0.7 entities for each utterance in the 3 domains respectively.
To evaluate model performance specifically on noisy user inputs,
we select queries from the test sets that are marked as 
containing speech transcription or user errors by the annotators and report results on this ``noisy" subset, which constitutes 13.5\%, 12.7\% data for movie\&TV and music domain respectively when an entity exists. \footnote{Sports domain does not have the annotation for noisy data available when this experiment was conducted.} 
To evaluate the relation feature, we also look at the ``related" subset where a valid relation exists in the utterance. This subset consists 13.4\% and 5.3\% of data for the music and sports domain with at least one entity. 
\footnote{Our KB does not have relation information for movie\&TV domain.}

We first train the NER model described in Section \ref{sec:ner}.
Next, for every example in our training dataset, we run inference on the trained NER model and generate the
top-5 NER hypotheses using beam search.
Following this, we retrieve the top 25 candidates for each of these hypotheses using our search engine 
combined with the ground truth NER and EL labels and fed to the EL model for training.

To measure the NER model performance, we use the standard NER F1 metric used for the CoNLL-2003 shared task \cite{tjong-kim-sang-de-meulder-2003-introduction}.
To measure the quality of the top-5 NER hypotheses, we compute the oracle top-5 F1 score by comparing and choosing 
the best alternative hypothesis among the $5$ and calculate its F1 score, for each test utterance. In this manner, we also know the upper 
bound that EL can reach from reranking NER hypotheses. 
As described in section \ref{subsec:el}, the EL model is optimized to perform two tasks simultaneously:
entity linking and reranking of NER hypotheses.
Hence to evaluate the performance of the EL model, we use two metrics: reranked NER-F1 score and the
EL-F1 score.
The reranked NER F1 score is measured on the NER predictions according to the top EL hypothesis, and is defined in the same way as the previous NER task.
To evaluate entity linking quality, we adopt a strict F1 metric similar to the one used for NER.
Besides entity boundary and entity type, the resolved entity also needs to be correct for the entity prediction
to be counted as a true positive.

\label{sec:hyperparams}

For NER model training, we use standard mini-batch gradient descent using the Adam optimizer with an initial
learning rate of 0.001, a scheduled learning rate decay of 0.99, LSTM with a hidden
layer of size 350 and a batch size of 256.
We apply a dropout of 0.5 to the embedding and biLSTM layers, and include token level
gazetteer features \cite{ratinov-roth-2009-design} to boost performance in recognizing common entities.
We linearly project these gazetteer features and concatenate the projection with the 200 dimensional word embeddings
and 100 dimensional character embeddings which are then fed into the biLSTM followed by the CRF.

For EL, the character CNN model we use has two layers, each with 100 convolution kernels of size 3, 4, and 5. Character embedding are 25 dimensional and trained end to end with the entity linking task. The MLP for embedding similarity takes the concatenation of two name embeddings, as well as their element-wise sum, difference, minimum, maximum, and multiplication. It has two hidden layers of size 1024 and 256, with output dimension 64. Similarity features of mentions in the prediction are averaged, while the other features like NER confidence and entity popularity are concatenated to the representation. The final MLP for scoring has two hidden layers, with size 256 and 64. We train the model on 4 GPUs with synchronous SGD, and for each gradient step we send a batch of 100 examples to each GPU. 

%% file: sections/s5-results.tex
\section{System Evaluation}
\label{sec:expt_res}

\input{sections/s5.1-ner_el.tex}
\input{sections/s5.2-case_study.tex}

\input{sections/s5.3-application.tex}

%% file: sections/s5.1-ner_el.tex
\subsection{Results}
\label{subsec:el-results}

We present F1 scores in different domains of the NER and EL model in Table \ref{tb:nen-best}. Since the EL model takes 5 NER hypotheses as input, it also acts as a re-ranker of the NER model, and we show substantial improvements on top-1 NER F1 score consistently over all test sets.

\begin{table}[h]
\begin{center}
\begin{tabular}{|@{\hskip3pt}l@{\hskip3pt}|r|c|l@{\hskip2pt}|}
\hline
            & \multicolumn{1}{c|}{\textbf{\begin{tabular}[c|]{@{}c@{}}NER F1\\ top-1/top-5\end{tabular}}} & \multicolumn{1}{c|}{\textbf{\begin{tabular}[c|]{@{}c@{}}reranked\\ NER F1\end{tabular}}} & \multicolumn{1}{@{\hskip4pt}l@{\hskip2pt}|}{\textbf{EL F1}} \\ \hline
\textbf{movie\&TV}      & 78.76 / 96.83                              & 81.62                                       & 79.67                            \\
\textbf{music} & 84.27 / 97.26                              & 87.40                                       & 84.95                            \\
\textbf{sports}      & 92.97 / 99.15                             & 93.48                                       & 91.13                           \\ \hline
\end{tabular}
\caption{Results for the best model setting. NER F1 are reported on the top-1 and top-5 NER prediction from the NER model that provides features for EL. Reranked NER F1 and EL F1 are reported on top-1 prediction from the best EL model selected by development sets.}
\label{tb:nen-best}
\end{center}
\end{table}

\begin{table*}[!htb]
    \begin{minipage}{.265\linewidth}
      \centering
      \begin{tabular}{|l|r|}
        \hline

	\hline
	\multicolumn{1}{|c|}{\textbf{}} & \multicolumn{1}{c|}{\textbf{\begin{tabular}[c]{@{}c@{}}{\bf (a)}\\+ MLP  \end{tabular}}} \\ \hline
	\textbf{movie\&TV}             & +3.58                                                                                     \\
	\textbf{(noisy)}     & +9.67                                                                                      \\ \hline
	\textbf{music}                  & +2.05                                                                                     \\
	\textbf{(noisy)}          & +10.03                                                                                      \\ \hline
	\end{tabular}
    \end{minipage}%
    \begin{minipage}{.27\linewidth}
      \centering
	\begin{tabular}{|l|r|}
	\hline
                                          &   \textbf{ (b)\, \, \, \,} \\
                                          & \textbf{+ Relation} \\ \hline
	\textbf{music}                  & +0.86                                     \\
	\textbf{(related)}        & +1.97                                     \\ \hline
	\textbf{sports}                 & +0.07                                     \\
	\textbf{(related)}       & +0.81                                     \\ \hline
	\end{tabular}
    \end{minipage} %
    \begin{minipage}{.4\linewidth}
      \centering
        \begin{tabular}{|l|r|r|}
	\hline
	\multicolumn{1}{|c|}{\textbf{}} & \multicolumn{1}{c|}{\textbf{\begin{tabular}[c]{@{}c@{}}{\bf (c)}\\+ Utterance \\ Embedding\end{tabular}}} & \multicolumn{1}{c|}{\textbf{\begin{tabular}[c]{@{}c@{}}{\bf (d)}\\+ Log-scale \\ Popularity\end{tabular}}} \\ \hline
	\textbf{movie\&TV}             & +0.25                                                                                           & +0.27                                                                                            \\
	\textbf{music}                  & +0.39                                                                                           & +0.02                                                                                            \\
	\textbf{sports}                 & -0.07                                                                                          & +0.08                                                                                            \\ \hline
	\end{tabular}
    \end{minipage} %
    \caption{EL mean F1 relative  \% improvements, reported on 10 runs average.}
    \label{tb:nen-improve}
\end{table*}

In Table \ref{tb:nen-improve}, we show improvements achieved by several specific model design choices and features on
entity linking performance. Table~\ref{tb:nen-improve}(a) shows the MLP similarity substantially improves entity linking accuracy with its capacity to
model text variations, especially on utterances with noisy entity mentions. The relation feature is powerful for disambiguating entities with similar names, and we show a considerable improvement in EL F1 on the subset of utterances that have related entities in Table~\ref{tb:nen-improve}(b). Table~\ref{tb:nen-improve}(c) shows utterance embeddings brought
improvements in the music, and media \& TV domains. The improvement brought by log-scale popularity feature is the largest for the movie \& TV domain as shown in Table~\ref{tb:nen-improve}(d), where the popularity distribution has extremely long tails compared to other
domains.

%% file: sections/s5.2-case_study.tex
\subsection{Qualitative Analysis}
\label{sec:qualitative}

We provide a few examples to showcase the effectiveness of our NEU system.
Firstly, the EL model is able to link noisy entity mentions to the corresponding entity canonical name in the knowledge
base. For instance, when the transcribed  utterance is ``play Carla Cabello'', the EL model is able to resolve the
mention ``Carla Carbello'' to the correct artist name ``Camila Cabello''.

Secondly, the EL model is able to recover from errors made by the NER system by leveraging the knowledge base to disambiguate entity mentions. The reranking is especially powerful when the utterance contains little context of the entity for the NER model to leverage. For example, for ``Doctor Strange'', the top NER hypothesis labels the full utterance as a generic ``Person'' type, and after reranking, EL model is able to leverage the popularity information (``Doctor Strange'' is a movie that was recently released and has a high popularity in our knowledge base) and correctly label the utterance as ``movieTitle''.
Reranking is also effective when the entity mentions are noisy, which will cause mismatches for the gazetteer features
that NER uses. For  ``play Avengers Age of Ultra'', the top NER hypothesis only predicts ``Avengers'' as ``movieTitle'', while after reranking, the EL model is able to recover the full span  ``Avengers Age of Ultra'' as a ``movieTitle'', and resolve it to  ``Avengers: Age of Ultron'', the correct canonical title.

The entity relations from the knowledge base are helpful for entity disambiguation. When the user refers to a sports
team with the name ``Giants'', they could be asking for either ``New York Giants'', a National Football League  (NFL) team, or ``San Francisco Giants'', a Major League Baseball  team. When there are multiple sports team mentions in an utterance, the EL model leverages a relation feature from the knowledge base indicating whether the teams are from the same sports league (as the user is more likely to mention two teams from the same league and the same sport). Knowing entity relations, the EL model is able to link the mention ``Giants'' in ``Cowboys versus Giants'' to the NFL team, knowing that ``Cowboys'' is referring to ``Dallas Cowboys''.

%% file: sections/s5.3-application.tex
\begin{table}[h]
    \begin{center}
    \begin{tabular}{|l|c|c|c|}
    \hline & A & B & C \\ 
    \hline
    \bf DC & 88.95 & 89.46 & \bf 90.04 \\
    \hline
    \hline
    \bf SP [movie\&TV] & 89.62 & 90.99 & \bf 91.67 \\
    \bf SP [music] & 83.97 & 84.26 & \bf 84.42 \\
    \bf SP [sports] & 86.37 & \bf 86.47 & 86.46 \\
    \hline
    \end{tabular}
    \end{center}
    \caption{\label{tab:ic_sp} Results for domain classifier (first row) and semantic parser. A is the baseline, B is
      A+NER, C is A+NER+EL.}
    \vspace*{-0.2in}
\end{table}

To validate the utility of our proposed NEU framework, we illustrate performance improvements in 
the Domain Classifier and the Semantic Parsers corresponding to the three domains (music, movies \& TV and sports) as described in Section \ref{subsec:method_application}. Table~\ref{tab:ic_sp} reports the classification accuracy for the Domain Classifier and the parse
accuracies for the Semantic Parsers (the model is said to have predicted the parse correctly if all the tokens are
tagged with their correct semantic parse labels). We observe substantial improvements in all 4 cases when NER features
are used as additional input, given all the other components of the system being the same. In turn, we observe further improvements when our NER+EL featurization is used.

%% file: sections/s6-conclusion.tex
\section{Conclusion}
\label{sec:concl}

In this work, we have proposed a Named Entity Understanding framework that jointly identifies and resolves entities present in an utterance when a user interacts with a voice assistant. Our proposed architecture consists of two modules: NER and EL, with the EL serving the additional task of possibly correcting the recognized entities from NER by leveraging rich signals from entity links in the knowledge base while simultaneously linking these entities to the knowledge base. With several design strategies in our system targeted towards noisy natural language utterances, we have shown that our framework is robust to speech transcription and user errors that occur frequently in spoken dialog systems. We have also shown that featurizing the output of NEU and feeding these features into other language understanding tasks substantially improves the accuracy of these models.

%% file: sections/s7-ethics.tex
\section{Ethical Considerations}
\label{sec:ethics}
We randomly sampled transcripts from Siri
production datasets over a period of months, and we believe it to be a representative sample
of usage in the domains described. In accordance with Apple’s privacy practices with respect to Siri requests, Siri
utterances are not associated with a user’s Apple ID, email address, or other data Apple may have from a user’s use of
other Apple services. In addition to Siri’s baseline privacy guarantees, we filtered the sampled utterances to remove
transcripts that were too long, contained rare words, or contained references to
contacts before providing the dataset to our annotators.

%% file: sections/sa-appendix.tex
\section{Architecture Diagram}
\label{app:arch_diag}

\begin{figure}[h]
  \centering
  \includegraphics[width=\linewidth]{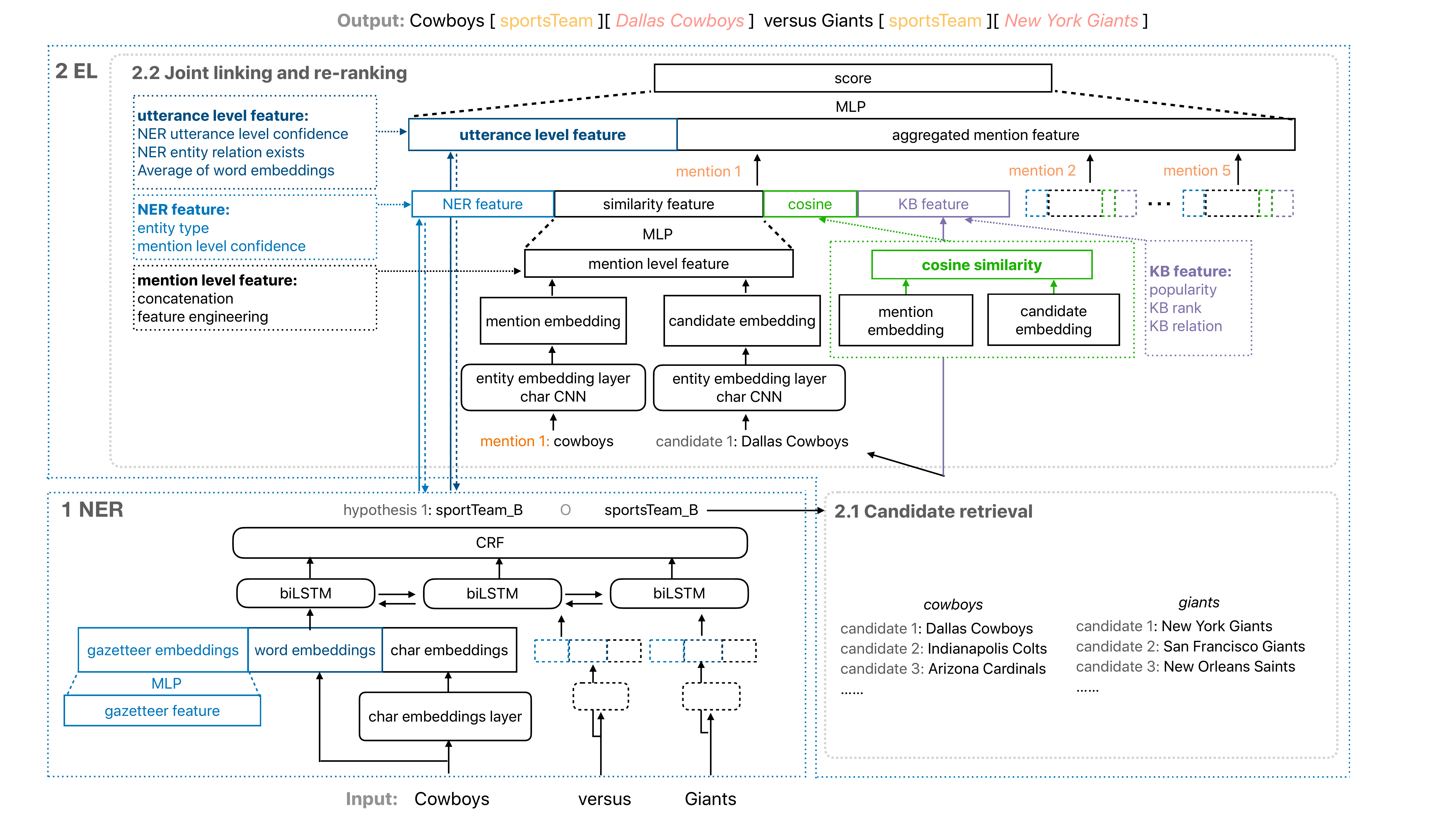}
  \caption{NEU system architecture.}
  \label{fig:neu_architecture}
\end{figure}